\documentclass{svproc}
\usepackage{graphicx}

\usepackage{amssymb}
\usepackage{amsmath}
\usepackage{array}
\usepackage{float}
\usepackage{xcolor}
\usepackage{url}

\begin{document}
\mainmatter              
\title{A Collaborative Robot-Assisted Manufacturing Assembly Process}
\titlerunning{Robot-assisted manufacturing process} 
%
\author{Miguel Neves\inst{1} \and Laura Duarte\inst{1} \and Pedro Neto\inst{1}}
\authorrunning{Miguel Neves et al.} 
%
\tocauthor{Miguel Neves, Laura Duarte, and Pedro Neto}
\institute{University of Coimbra, CEMMPRE, ARISE, Department of Mechanical Engineering, 3030-788, Coimbra, Portugal\\
\email{pedro.neto@dem.uc.pt}}

\maketitle              

\begin{abstract}
An effective human-robot collaborative process results in the reduction of the operator’s workload, promoting a more efficient, productive, safer and less error-prone working environment. However, the implementation of collaborative robots in industry is still challenging. In this work, we compare manual and robot-assisted assembly processes to evaluate the effectiveness of collaborative robots while featuring different modes of operation (coexistence, cooperation and collaboration). Results indicate an improvement in ergonomic conditions and ease of execution without substantially compromising assembly time. Furthermore, the robot is intuitive to use and guides the user on the proper sequencing of the process. 

\end{abstract}


\section{Introduction}
\label{sec:1}

Manufacturing assembly often requires the human operator to conduct complex tasks in non-ergonomic conditions. An effective human-robot interactive and collaborative process stands out among the possible solutions to reduce human workload and improve comfort, by having a robot to assist/answer the human needs in the manufacturing context \cite{Lorenzini2023}. This approach has the main advantage of combining the flexibility and dexterity of human operators with the consistency and precision of collaborative robots. 

The human-robot interactive process can be subdivided into three main types, human-robot coexistence, human-robot cooperation and human-robot collaboration \cite{Gjeldum2022,Kopp2020}. Human-robot coexistence, as the name suggests, occurs when the human operator and the robot share a physical space but do not share a workspace. However, since there is no physical separation between them, the robot must be equipped with safety systems that stop the robot when a human operator enters its workspace. Human-robot cooperation refers to the scenario where the human and robot share a workspace but their tasks are organized in a sequential way and the physical contact between the two is not required. Lastly, human-robot collaboration corresponds to the scenario where the human operator and the robot share a workspace, work simultaneously on the same task and physical contact often occurs \cite{Saenz18}. The ISO/TS 15066 standard extends the generic guidelines of the ISO 10218 standard of safety requirements for industrial robots by defining passive and active safeguards that robot designers, integrators and users must acknowledge when deploying collaborative robot applications. \cite{Chemweno20}

Human-robot collaborative scenarios are complex to implement due to the requirement of being adaptable and safe for the human operator \cite{Malik2020}. Due to these difficulties, frameworks have been proposed to aid the implementation processes \cite{Maurice2017} and the related task allocation processes \cite{Gjeldum2022}. Human-robot interaction and collaboration must be dynamic to accommodate the current workflow of the human operator, reducing the workload felt by the human operator \cite{Oliff2020}. In fact, the human operator's state and mental workload are some of the most important aspects to be considered in human-robot interactive and collaborative scenarios. As a result, various studies have focused on capturing this information through the usage of wearable sensors coupled with machine-learning solutions \cite{Lin2022,Buerkle2022}. Alternatively, virtual environments offer a safe space to verify the plausibility of the real-world implementation, facilitating the realisation of an integrated, flexible and collaborative manufacturing environment \cite{Malik2020,Onaji2022}. These environments can also be used to add useful and relevant information to the real-world environment through techniques such as augmented reality (AR) \cite{Aivaliotis2023}.

In this work, inspired by a real industrial application, we propose to use a collaborative robot as a third hand to assist/help a human operator in a manufacturing assembly process. The proposed methodology aims to identify opportunities for the collaborative robot to assist the operator in the various assembly tasks. Finally, this human-robot collaborative process is compared with the existing manual process, highlighting and discussing the pros and cons of each modality.

\begin{figure*}[t]
    \centering\includegraphics[width=0.97\textwidth]{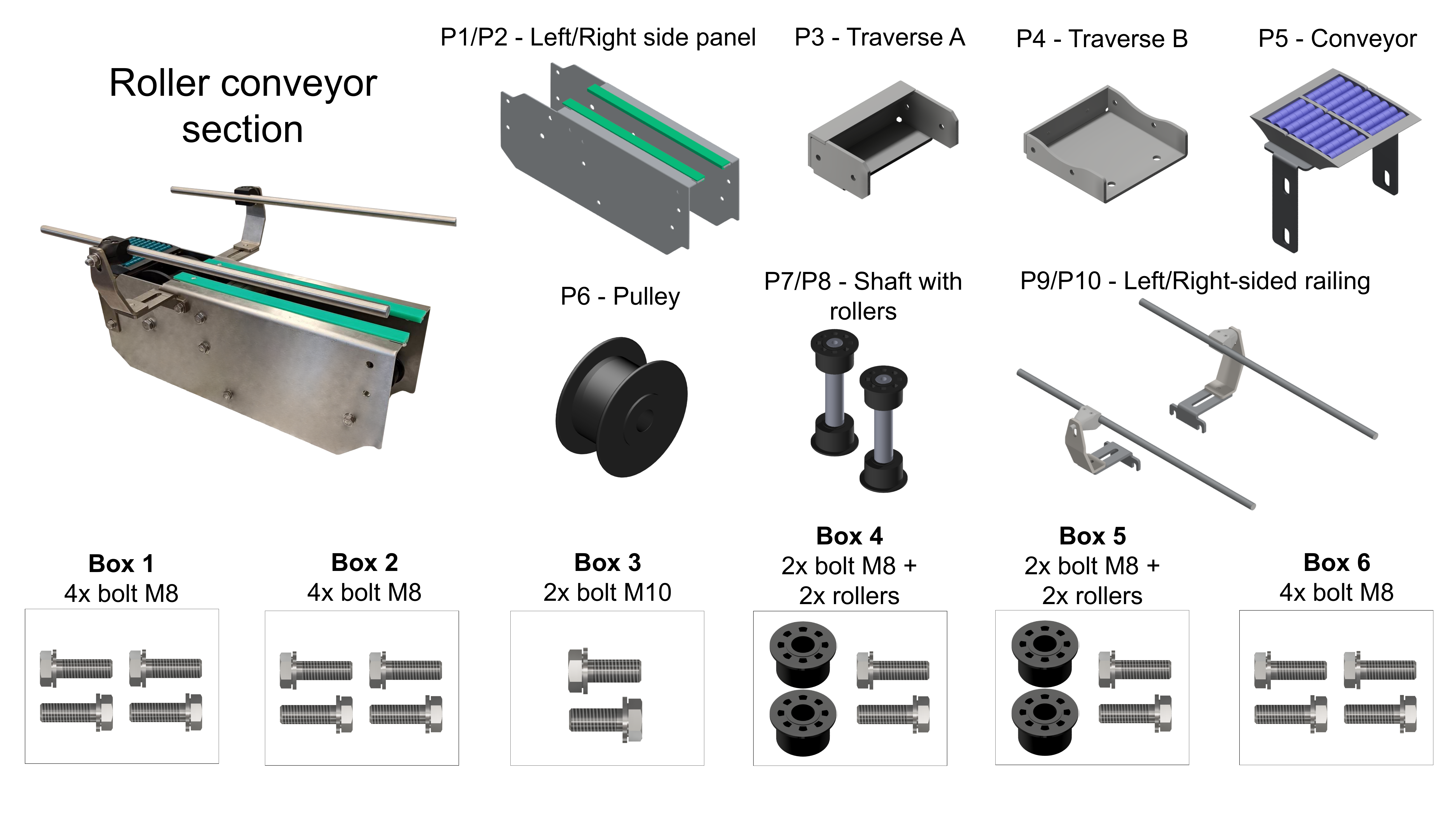}
    \caption{Components from the roller conveyor section, including the contents of the boxes transported by the robot to the operator.}
    \label{fig:Setup}
\end{figure*}


\section{Methodology} 
\label{sec:2}

\begin{figure*}[t]
    \centering\includegraphics[width=\textwidth]{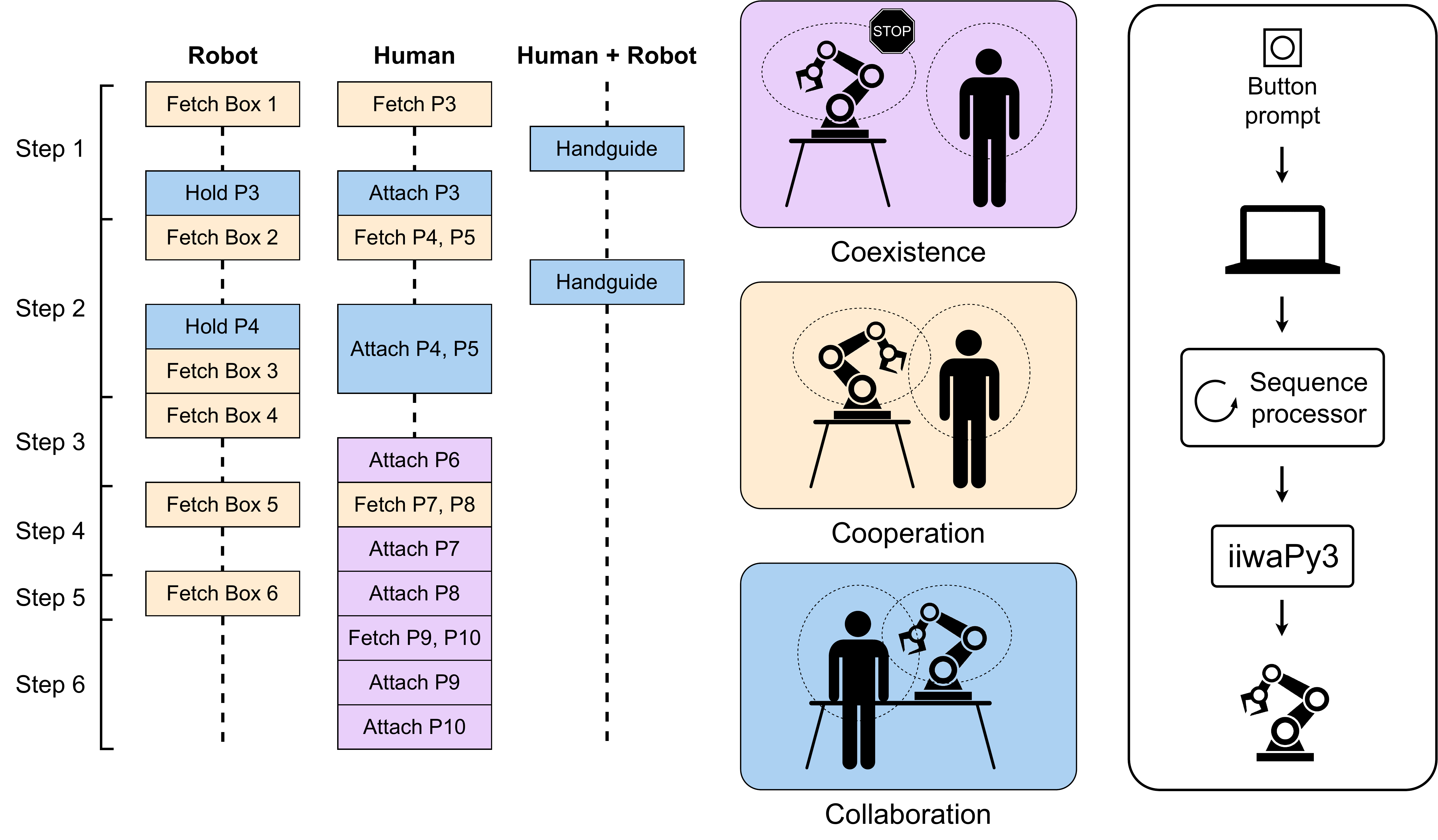}
    \caption{Sequence of actions done by the human, the robot or in collaboration in the collaborative scenario.}
    \label{fig:Sequence}
\end{figure*}

This study considers the real industrial use case of the assembly of a roller conveyor section, Fig~\ref{fig:Setup}. It has a total of 10 parts (P1 to P10) plus the components delivered in boxes (Box 1 to Box 6), and the assembly is subdivided into 6 steps. The process is studied considering two different scenarios, namely the manual assembly and the human-robot collaborative assembly to identify whether the addition of a collaborative robot can positively impact the assembly task. The metrics considered are the ergonomic conditions, the correctness of the assembly sequence and the execution time. In the first scenario, manual assembly, all bolts and components are within reach of the human operator. In the second scenario, the collaborative robot is introduced with two major functions: bringing boxes with small components and bolts to the workspace and holding components during the assembly, operating as a third hand to the human operator. In this scenario, bigger components remain within reach of the human operator, Fig~\ref{fig:Sequence}. 

The collaborative robot is a KUKA lbr iiwa 7 r800 \cite{KUKAiiwa} equipped with a SCHUNK Co-act EGP-C 64 gripper \cite{SCHUNKgripper}. The human operator has an input button, which prompts the pre-programmed sequence processor to send the next sequence of commands to the collaborative robot through the iiwaPy3 library \cite{Safeea2023}. Certain sequences require the human operator to guide the collaborative robot into position through an intuitive process of hand-guiding at the end-effector level \cite{Safeea2022}. For each assembly step, we defined the tasks being performed according to the current process of manual assembly in the company, and the associated collaborative mode when the robot is used. The coexistence refers to physical space sharing, but not workspace sharing, while the human is performing assembly actions. Cooperation refers to when the robot brings the boxes of components to a specific area in a given sequence, while the human is performing assembly actions within the workpiece. The collaboration corresponds to the hand-guiding of the robot to specific areas to hold components while the operator is performing assembly actions such as screwing. In this case, the robot acts as a third hand to the operator.
Ergonomy levels can be differentiated into three categories: Undesirable if the user experiences physical discomfort due to unnatural body positioning or handling heavy loads; average if the user experiences some discomfort but it does not impact their performance; and good if the user experiences minimal effort.


\section{Results and discussion}
\label{sec:3}
\vspace*{-3pt}

\begin{figure*}[t]
    \centering\includegraphics[width=0.99\textwidth]{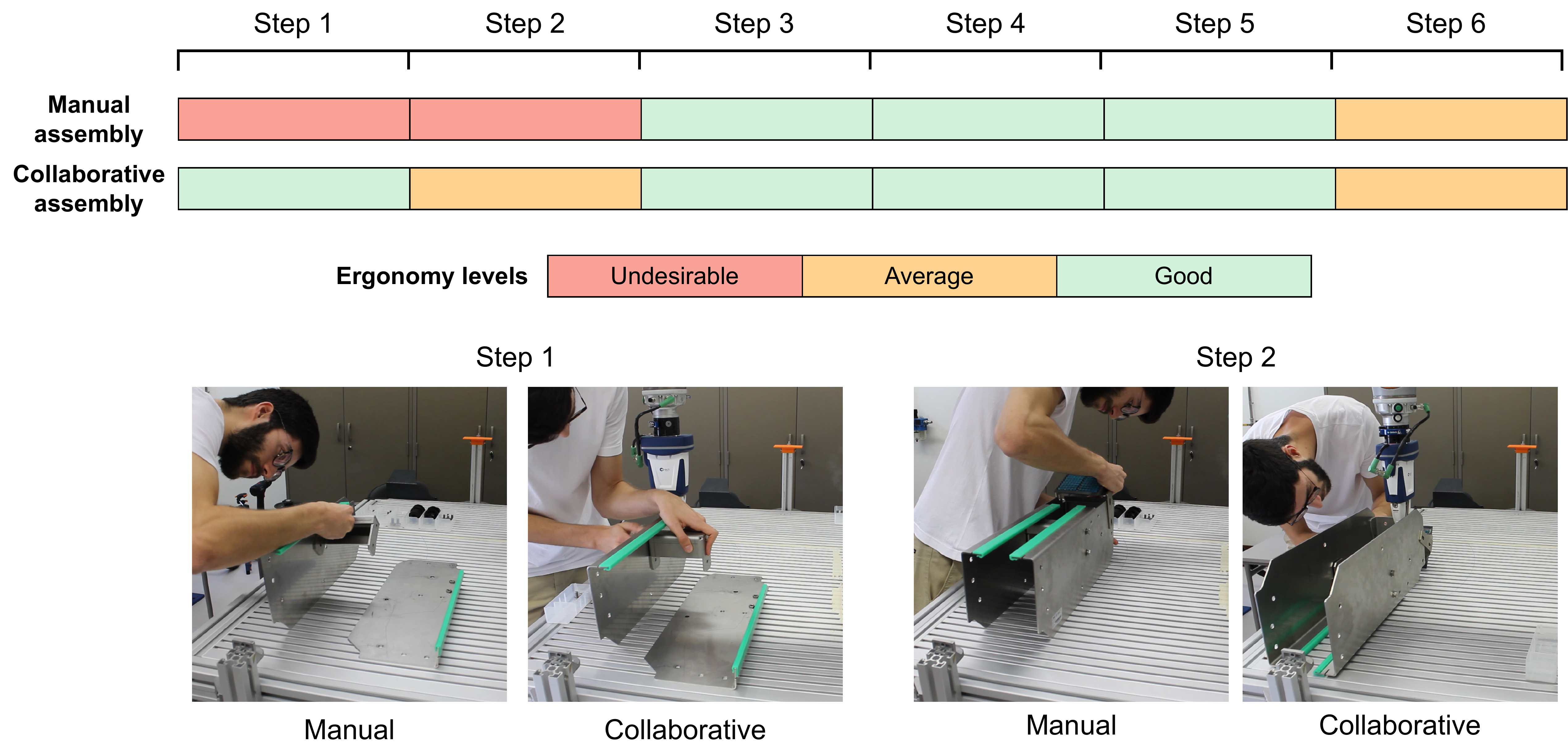}
    \caption{Ergonomics comparison between manual and collaborative assemblies for each different assembly step.}
    \label{fig:Ergonomics}
\end{figure*}

Results indicate that for this specific part, the measured manual assembly time was about 6 minutes, while the collaborative assembly took approximately 7 minutes. This equates to a $15\%$ increase in task time duration when using the human-robot collaborative approach. However, the benefits gained from using the collaborative robot outweigh the savings in manual assembly time. The human operator struggles in the manual assembly tasks which require both hands, particularly steps 1 and 2, Fig.~\ref{fig:Ergonomics}. Notably, step 2 demands the operator to hold two components while simultaneously placing a screw to hold them together. The robot acting as a third hand can be used to hold one of the components, facilitating the assembly process. As a consequence, the introduction of the collaborative robot simplifies the execution of assembly steps 1 and 2. Most assembly tasks benefit from collaborative assistance, even when it is not essential to the assembly process. By assigning the robot the task of holding certain components in place, the operator can be in a more ergonomic position and, thus, perform the assembly task with more ease. The human operator also benefits from a more intuitive workflow. The robot interaction is triggered by a single button, which is selected depending on the current step of the assembly process.

Owing to the diversity of the components used in the assembly process, a variety of screw sizes are needed to join them. Therefore, the robot is able to deliver the correct screws for each step of the assembly process, reducing the chance of errors by the human operator and clearing up the workspace. Due to the low number of trials performed, it was impossible to make a thorough quantitative analysis of the error reduction in this regard, but it was possible to conclude that the mental workload was lower in the collaborative scenario.

\section{Conclusions}
\label{sec:4}

This study demonstrates the usage of a robot as a collaborative assistant to a human operator in a manufacturing assembly process. Results demonstrated that collaborative robots can be introduced in an intuitive way in the assembly process, as the human operator only has the additional basic and intuitive tasks of positioning the robot through hand-guiding and prompting the robot to continue the sequence using a single button.

By using a collaborative robot, the human operator benefits from improved ergonomic conditions, a well-organized workspace, and less mistakes in the assembly sequencing. The use of the robot as a third hand to hold components facilitates the most strenuous assembly tasks. However, the total assembly time increases $15\%$ when compared to the manual assembly. 

While the proposed assembly process covers some of the main primitive tasks in assembly future work will be dedicated to testing with other components, namely from electronics and textile industries.


\subsubsection*{Acknowledgements}
This research is sponsored by national funds through FCT – Fundação para a Ciência e a Tecnologia, under the project UIDB/00285/2020 and LA/P/0112/2020, and the grants 2021.06508.BD and 2021.08012.BD.


\end{document}